\documentclass{article}
\usepackage{spconf,amsmath,graphicx,hyperref}
\usepackage{mathbbol}
\usepackage[table,xcdraw]{xcolor}
\usepackage{bbding}

\title{Prototype-Based Pseudo-Label Denoising for Source-Free Domain Adaptation in Remote Sensing Semantic Segmentation}
\name{
	Bin Wang$^{\dagger}$ \qquad 
	Fei Deng$^{\ddagger}$ \qquad 
	Zeyu Chen$^{\dagger}$ \qquad 
	Zhicheng Yu$^{\dagger}$ \qquad 
	Yiguang Liu$^{\dagger}$ \textsuperscript{\Envelope} \qquad 
	\thanks{This work is supported by the National Key R\&D Program of China 
		(No.2023YFF0615800), Sichuan Science and Technology Program under Grant 2024ZHCG0191.}
	\thanks{\textsuperscript{\Envelope} Yiguang Liu are the corresponding authors. Email: liuyg@scu.edu.cn.}
	}
\address{
	$^{\dagger}$ College of Computer Science, Sichuan University \\ 
	$^{\ddagger}$ College of Computer Science and Cyber Security, Chengdu University of Technology
}
\begin{document}
%
\maketitle
\begin{abstract}

Source-Free Domain Adaptation (SFDA) enables domain adaptation for semantic segmentation of Remote Sensing Images (RSIs) using only a well-trained source model and unlabeled target domain data. However, the lack of ground-truth labels in the target domain often leads to the generation of noisy pseudo-labels.
Such noise impedes the effective mitigation of domain shift (DS).
To address this challenge, we propose ProSFDA, a prototype-guided SFDA framework. It employs prototype-weighted pseudo-labels to facilitate reliable self-training (ST) under pseudo-labels noise. We, in addition, introduce a prototype-contrast strategy that encourages the aggregation of features belonging to the same class, enabling the model to learn discriminative target domain representations without relying on ground-truth supervision. Extensive experiments show that our approach substantially outperforms existing methods. The source code is publicly available at \url{https://github.com/woldier/pro-sfda}.

\end{abstract}
\section{Introduction}
\label{sec:intro}


In recent years, deep learning-based methods have achieved significant progress in this task, particularly approaches based on Fully Convolutional Networks (FCNs) \cite{long2015,Chen2018} and Transformer-based architectures \cite{Zheng2021,Xie2021}. However, deep neural networks remain highly susceptible to DS between source and target data captured under different conditions \cite{Chen2021}.

To address the challenge of domain shift (DS), Unsupervised Domain Adaptation (UDA) methods \cite{Toldo2020,gao2023,wang2024} have emerged as a prominent research direction. However, these approaches typically require access to labeled source domain RSIs during training, which imposes certain limitations. 
In scenarios where source data cannot be obtained, conventional UDA methods become difficult to apply.

To address this issue, SFDA has been proposed, in which only a pre-trained source model and unlabeled target domain data are available. Under this setting, SFDA offers broader applicability than traditional UDA methods, as it performs domain adaptation solely within the target domain without relying on source data.

Recently, numerous SFDA methods  \cite{Chen2021,liu2021,you2021,kundu2021,cao2024,gao2024} for semantic segmentation have been proposed, which can generally be categorized into two groups: generative-based and self-training-based approaches. Generative-based SFDA methods \cite{liu2021, kundu2021} aim to synthesize source-like data to reduce the DG between the source and target domains. While these methods can achieve promising results, generating realistic and source-consistent samples remains challenging and often incurs significant computational overhead.
In contrast, ST-based SFDA methods \cite{you2021,Akkaya2022,cao2024,gao2024} adopt a more practical and constrained setting, relying solely on a well-trained source model and unlabeled target data. These methods adapt the model to the target domain through pseudo-labeling. 

In real-world RSIs applications, model adaptation commonly involves DS due to differences across geographic regions or sensor types. Other factors—such as object shape variations across domains—often contribute less significantly to DS in this context. Given this, a portion of the model’s predictions on target data is likely to be reasonable and correct, providing a useful foundation for pseudo-label-based ST. 
However, due to inherent DS, some prediction errors are unavoidable, leading to noisy pseudo-labels that can hinder the self-adaptation process. Therefore, effectively reducing the impact of noisy pseudo-labels is critical for achieving optimal adaptation performance.

In this paper, we propose a novel method, ProSFDA, for SFDA in RSIs semantic segmentation. 
To mitigate the negative impact of noisy pseudo-labels during training, we introduce a prototype-weighted pseudo-labeling mechanism.
This strategy provides a more reliable supervision signal by assigning higher confidence to predictions aligned with class prototypes. These prototypes are initially computed by passing target data through the source model and are subsequently updated via exponential moving average (EMA) using predictions from the evolving target model. To further improve pseudo-label stability, we employ an EMA-based teacher model to generate less perturbed pseudo-labels.
However, denoising pseudo-labels may limit the model’s ability to learn from uncertain but informative target data. To address this, we incorporate a prototype-contrast strategy that encourages clustering of features belonging to the same class, thereby enhancing representation learning from the target domain. This strategy compensates for the loss of learning due to noise suppression and strengthens the model's alignment with the target task.
In summary, our contributions are as follows:

(1) We propose ProSFDA, a SFDA framework tailored for RSI semantic segmentation, which adapts a well-trained source model to the target domain without requiring access to source data or target labels.

(2) We introduce a prototype-weighted pseudo-label denoising strategy to suppress the negative effects of noisy pseudo-labels, thereby improving adaptation under DS.

(3) We incorporate prototype-contrast feature clustering to aggregate semantically similar features, enabling more comprehensive learning from target data and enhancing task-specific adaptation.

\section{Methods}
\label{sec:method}

\subsection{Preliminary}

In the setting of Source-Free Domain Adaptation for semantic segmentation, the objective is to adapt a well-trained source model $M_S$, originally trained on the labeled source domain
$\mathcal{D}_S = \left\{ \left(x_S^i, y_S^i\right) \mid x_S^i ,\ y_S^i \right\}_{i=0}^{N_S - 1},$
to an unlabeled target domain
$\mathcal{D}_T = \left\{ x_T^i \mid x_T^i \right\}_{i=0}^{N_T - 1}.$
The goal is to enable model $g_\theta$, adapt form $M_S$, to perform accurate semantic segmentation on $\mathcal{D}_T$ without access to any source data or target labels during adaptation.

Ideal model adaptation to the target distribution typically requires access to ground-truth (GT) labels. However, this assumption does not hold in the SFDA setting. To address this challenge, a common approach is to assign pseudo-labels to the unlabeled target data, enabling model adaptation via ST.

\subsection{Generation of Pseudo Labels for the Target Domain}
\label{sec:method-pl}
Pseudo-labels can be generated using either offline \cite{gao2024, cao2024} or online \cite{Tranheden2021,wang2024} strategies . 
Existing SFDA methods primarily adopt offline pseudo-labeling, which results in stale supervision, as the labels do not evolve alongside model updates. In contrast, the online strategy dynamically updates pseudo-labels during training, which is especially beneficial in our framework since the use of prototypes requires features that better reflect the current state of the model. Therefore, we adopt an online pseudo-labeling strategy in this work.

In the online setting, pseudo-labels for the target data are generated by a teacher network $t_\theta$, which is updated during training alongside the student model $g_\theta$. Specifically, for a target image $x_T^i$, the pseudo-label $p_T^{i}$ at pixel location $(h, w)$ is defined as:
\begin{equation}
	p_{T}^{i,h,w} = \left [ c = \arg\max_c\ t_\theta(x_T^i)^{h,w,c} \right ],
\end{equation}
where $t_\theta(x_T^i)^{h,w,c}$ denotes the logit prediction for class $c$ at pixel $(h, w)$. The Iverson bracket $[\,\cdot\,]$ indicates the selection of the class with the highest confidence.
The teacher network $t_\theta$ is updated using an EMA of the student model’s parameters, ensuring stable and progressively refined supervision during training.

\begin{figure}[]
	\centering
	\includegraphics[width=\linewidth]{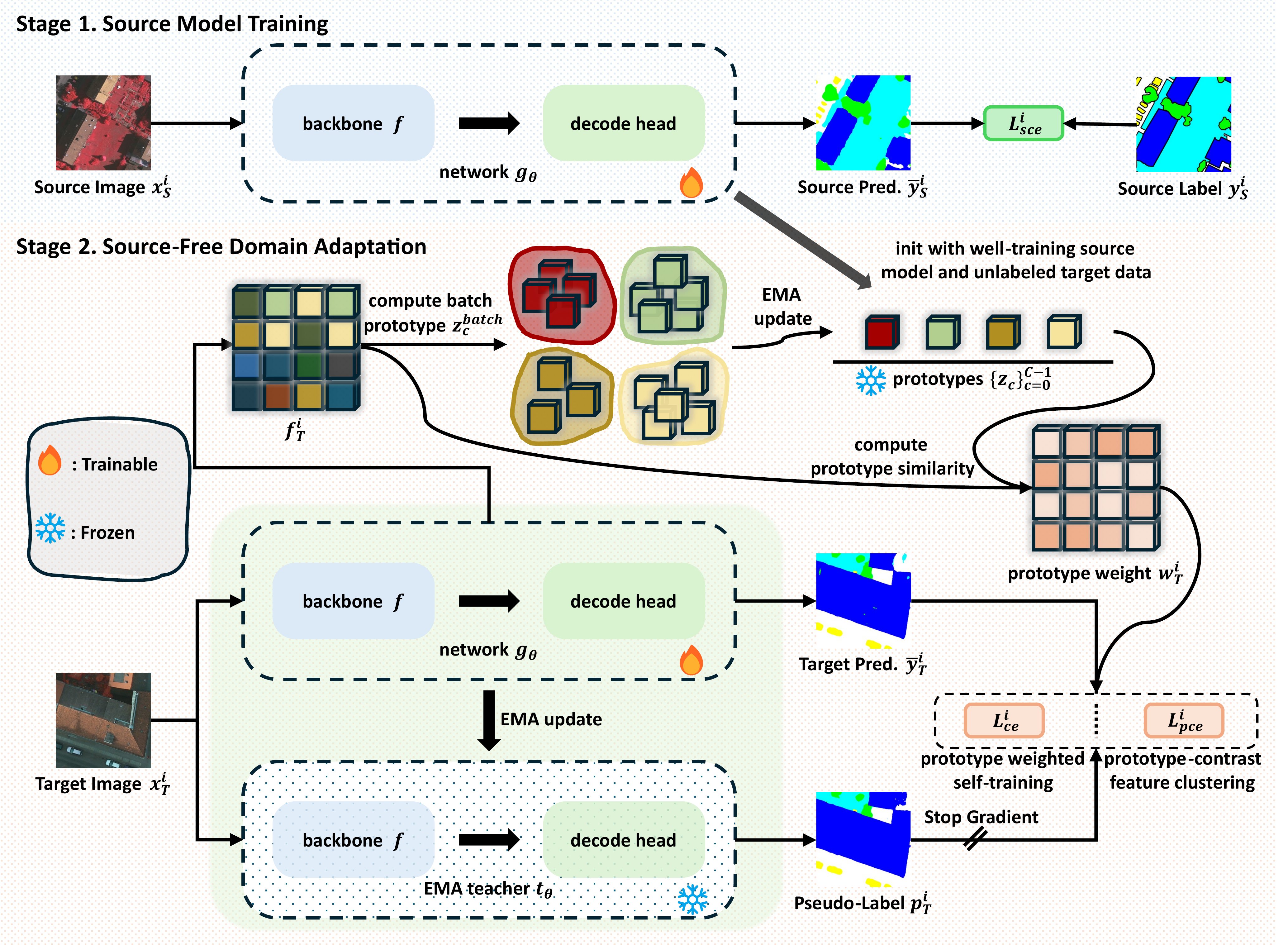}
	\caption{Network Architecture. Stage 1 illustrates the supervised training of the network on labeled source data. Stage 2 presents the proposed SFDA method, ProSFDA. 
		}
	\label{network}
\end{figure}

\subsection{prototype weighted self-training}

However, the pseudo-labels generated in Section \ref{sec:method-pl} may contain noise. Previous methods typically address this issue by filtering out unreliable pixel labels during pseudo-label generation, effectively discarding supervision for those regions. In contrast, we propose a different strategy. Instead of removing potentially noisy labels, we compute the similarity between each pixel's feature and the class prototypes. These similarity scores are then used as confidence weights to modulate the ST loss. Pixels with lower similarity receive smaller weights, thereby reducing the impact of noisy pseudo-labels on model adaptation.

At initialization, we use the pre-trained source model $M_S$ to initialize both the student $g_\theta$ and the teacher $t_\theta$. Given the target image $x_T^i$, we obtain the corresponding features $f_T^i$ and pseudo-labels $p_T^i$. The prototype for class $c$, denoted as $z^c$, is initialized as follows:
\begin{equation}
	z^c = \frac{\sum_{i,h,w} p_T^{i,h,w,c} \cdot f_T^{i,h,w}}{\sum_{i,h,w} p_T^{i,h,w,c}} 
\end{equation}

After initialization, the prototypes are updated during training via EMA, where $z_{\text{batch}}^c$ denotes the class-wise prototype computed from the current mini-batch:

\begin{equation}
	z^c \leftarrow \alpha \cdot z^c + (1 - \alpha) \cdot z_{\text{batch}}^c 
\end{equation}

Next, we compute the cosine similarity between the feature $f_T^{i,h,w}$ and each class prototype $z^c$, which serves as a confidence weight $w_T^{i,h,w,c}$:

\begin{equation}
w_T^{i,h,w,c} = \frac{f_T^{i,h,w} \cdot z^c}{\|f_T^{i,h,w}\| \cdot \|z^c\|} 
\end{equation}

The final pseudo-label used for ST supervision, denoted as $\hat{y}_p^i(h,w)$, is assigned based on the class with the highest similarity:

\begin{equation}
\hat{y}_p^{i,h,w} = \arg\max_{c'} w_T^{i,h,w,c'} 
\end{equation}

Finally, we optimize the model using a weighted cross-entropy loss, where the prototype-based confidence $w_T^i$ modulates the ST supervision:

\begin{equation}
\mathcal{L}_{\text{ce}}^i = -\sum_{h=0}^{H-1} \sum_{w=0}^{W-1} \sum_{c=0}^{C-1} w_T^{i,h,w,c}  \cdot p_T^{i,h,w,c} \cdot \log g_\theta(x_T^i)^{h,w,c} 
\end{equation}

\subsection{prototype-contrast feature clustering}

While ST with prototype-weighted pseudo-labels denoising effectively mitigates the adverse effects of noisy pseudo-labels in addressing DS, it has a notable limitation: low-confidence (i.e., noisy) pixels receive smaller weights, resulting in weaker supervision. Consequently, the model may fail to sufficiently learn from these pixels, particularly those affected by label noise. To address this issue, we propose prototype-contrast feature clustering, which aggregates similar features to enhance the model’s ability to learn from target data.

We introduce a prototype-enhanced cross-entropy loss, which leverages prototype-based reference predictions $\hat{y}_p$ to further improve the model's performance, especially on noisy classes:

\begin{equation}
\mathcal{L}_{pce}^i = -\sum_{h=0}^{H-1} \sum_{w=0}^{W-1} \sum_{c=0}^{C-1} \hat{y}_p^i(h,w,c) \cdot \log w_T^i(h,w,c) 
\end{equation}

Interestingly, we observe that the pseudo-labels $P_T^i$ generated by the teacher model and the prototype-based labels $\hat{y}_p^i$ may differ, introducing label ambiguity. This raises an important question: which label should be trusted more for supervision?

To address this, we propose a confidence-based resolution strategy. For a given pixel, we define confidence as the ratio between the top-1 and top-2 predicted probabilities. A higher ratio indicates a more confident prediction. Specifically, the confidence of a prediction from the teacher model and the prototype-based method is computed as:

\begin{equation}
	 C_{p_T}^{i,h,w} = \frac{\text{Rank}_1(g_\theta(x_T^i)^{h,w})}{\text{Rank}_2(g_\theta(x_T^i)^{h,w})}, \quad  
	 C_{\hat{y}_p}^{i,h,w} = \frac{\text{Rank}_1(w_T^{i,h,w})}{\text{Rank}_2(w_T^{i,h,w})} 
\end{equation}

where $\text{Rank}_j(\cdot)$ denotes the $j$-th highest value in the corresponding class probability vector.

Using this confidence-aware selection mechanism, we redefine the prototype-enhanced loss function as follows:

\begin{equation}
	\begin{aligned}
 \mathcal{L}_{\text{pce}}^i  & = -\sum_{h=0}^{H-1} \sum_{w=0}^{W-1} \sum_{c=0}^{C-1}  
 \left [p_T^i = \hat{y}_p^i \right ] \cdot \hat{y}_p^{i,h,w,c} \cdot \log w_T^{i,h,w,c} \\
 & + \left [ p_T^i \neq \hat{y}_p^i, C_{p_T}^i > C_{\hat{y}_p}^i \right] \cdot 	p_T^{i,h,w,c} \cdot \log w_T^{i,h,w,c} \\ 
 & + \left [p_T^i \neq \hat{y}_p^i, C_{p_T}^i < C_{\hat{y}_p}^i \right ] \cdot \hat{y}_p^{i,h,w,c} \cdot \log w_T^{i,h,w,c}
	\end{aligned}
\end{equation}
%

This confidence-guided loss function effectively enhances the supervision signal for noisy regions and improves the model’s learning capability on the target domain.

\section{EXPERIMENTAL RESULTS}
\begin{table*}[]
	\caption{ POT IRRG → VAI IRRG. The evaluation metrics used are IoU. All values are presented as percentages (\%).}
	\label{tab:my-table}
	\begin{tabular}{ccccccccc}
		\hline
		Method     & S.F. & Clutter & Car   & Tree  & Low Vegetation & Building & Impervious Surface & Overall \\ \hline
		DeepLabV3+ \cite{Chen2018} &  \XSolidBrush    & 3.78    & 17.79 & 66.01 & 30.82          & 69.37    & 41.93              & 38.28   \\
		Segformer  \cite{Xie2021} &  \XSolidBrush    & 5.73    & 25.04 & 72.13 & 50.41          & 76.54    & 60.45              & 48.39   \\
		LD    \cite{you2021}     &  \CheckmarkBold    & 37.41   & 8.61  & 65.51 & 50.70          & 73.84    & 70.68              & 51.13   \\
		IAPC    \cite{cao2024}   &  \CheckmarkBold    & 37.41   & 31.06 & 63.97 & 47.56          & 71.86    & 66.29              & 53.01   \\
		SFDA-DPL  \cite{Chen2021} &  \CheckmarkBold    & 17.34   & 44.84 & 72.80 & 54.23          & 80.64    & 66.73              & 56.10   \\
		Gao et al. \cite{gao2024} &  \CheckmarkBold    & 19.33   & 52.02 & 73.11 & 54.72          & 82.44    & 68.82              & 58.51   \\
		\rowcolor[HTML]{C0C0C0} 
		ProSFDA    &  \CheckmarkBold    & 33.01   & 60.34 & 72.65 & 52.51          & 83.59    & 75.30              & 62.90   \\ \hline
	\end{tabular}
\end{table*}
\subsection{Dataset}
To evaluate the effectiveness of the proposed method on cross-domain RSIs segmentation tasks, we employ two benchmark datasets: Potsdam and Vaihingen.
The Potsdam (POT) dataset consists of 38 high-resolution RS images.
 It offers three different image modalities: IR-R-G, R-G-B, and R-G-B-IR. In this study, we utilize the IR-R-G modality. The Vaihingen (VAI) dataset includes 33 RS images. 
 Only one image modality IR-R-G is available in the VAI dataset.

Potsdam primarily covers urban areas with a higher density of vehicles, whereas Vaihingen contains more residential buildings and farmland. Although both datasets share the same task objectives and label space, they exhibit significant differences in feature distributions, making them well-suited for evaluating the performance of SFDA methods. Based on this, we design one segmentation task:

Potsdam IR-R-G → Vaihingen IR-R-G (POT IRRG → VAI IRRG).

\subsection{Implementation Details}
Given the outstanding performance of Transformer-based architectures in semantic segmentation, we adopt Mix Transformers (MiT) \cite{Xie2021}, specifically designed for segmentation tasks, as the feature extraction backbone $f$. For the decoder head, we utilize a context-aware multi-scale feature fusion design proposed by Hoyer et al. \cite{Hoyer2022}. The network $g_\theta$ is optimized using the AdamW optimizer with hyperparameters set to $\beta = (0.9, 0.999)$, a weight decay of 0.01, and a learning rate of $6 \times 10^{-5}$. The update coefficient $\alpha$ for both the EMA teacher and prototype update is set to 0.99. All models are trained using 2 × Nvidia A100 GPUs to enhance training efficiency and performance.

\subsection{Comparison for Semantic Segmentation}
In this study, we compare our proposed method with several state-of-the-art SFDA approaches, including LD \cite{you2021}, IAPC \cite{cao2024}, SFDA-DPL \cite{Chen2021}, and the method proposed by Gao et al. \cite{gao2024}. To better evaluate the performance improvements brought by SFDA techniques, we also include source-only baselines—DeepLabV3+ \cite{Chen2018} and Segformer \cite{ Xie2021}.

\subsubsection{Evaluation of Performance}
For a fair comparison, we adopt the hyperparameter settings recommended in the original publications to ensure each method performs optimally. The quantitative results of these methods are presented in Table \ref{tab:my-table}. 

As evident from the results, the source-only methods struggle to effectively transfer knowledge from the source to the target domain, resulting in suboptimal segmentation performance. In contrast, the SFDA methods substantially improve segmentation accuracy.
ProSFDA achieves the best segmentation performance among all compared methods.

\begin{figure}[]
	\centering
	\includegraphics[width=\linewidth]{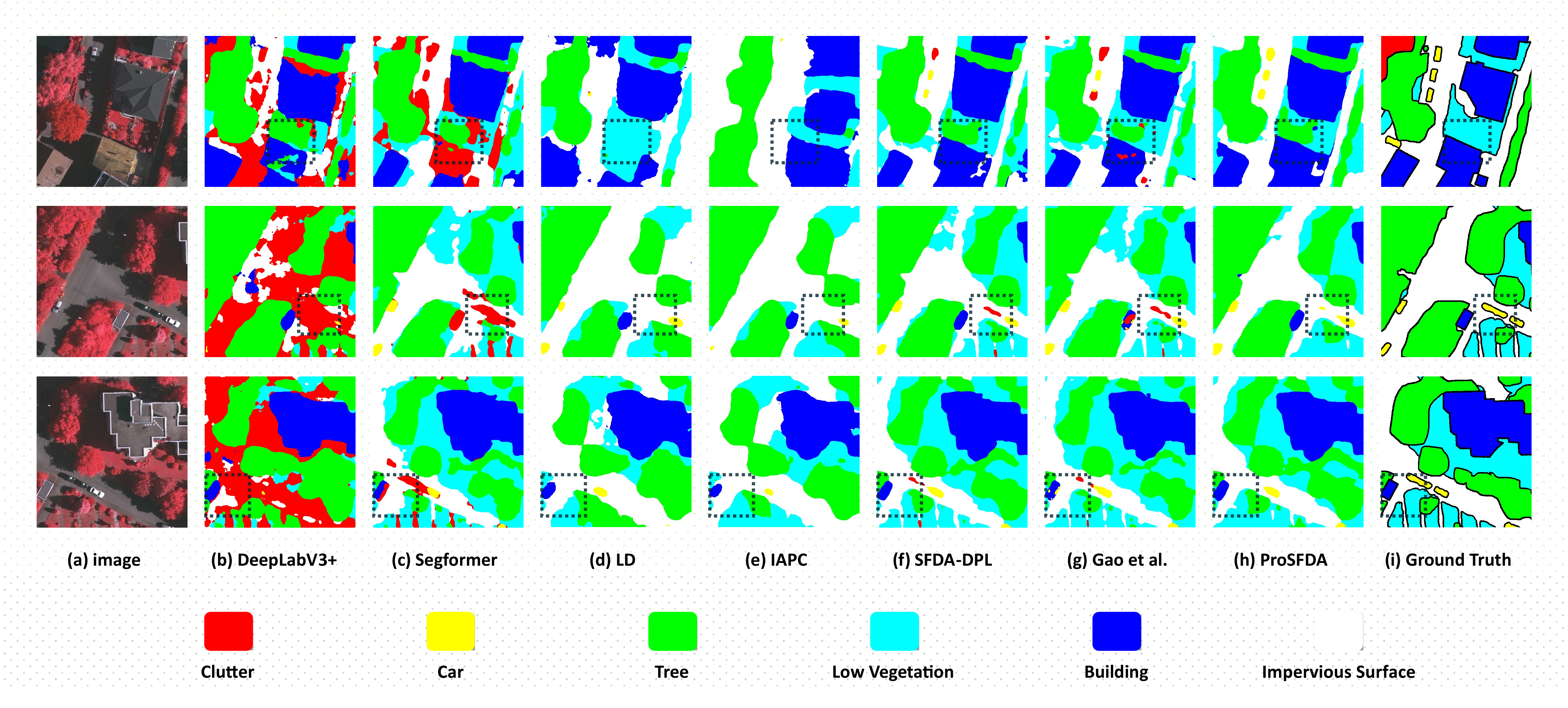}
	\caption{Visualization of results on POT IRRG → VAI IRRG}
	\label{visualization}
\end{figure}

\subsubsection{Visualization of Semantic Segmentation}
Fig. \ref{visualization} presents a comparison between the proposed ProSFDA and other existing methods. Overall, the experimental results are consistent with the previously reported quantitative findings. 
The proposed ProSFDA method introduces a prototype-weighted ST strategy to mitigate the influence of noisy pseudo-labels. Additionally, by incorporating prototype-contrastive feature clustering, it effectively learns discriminative representations from the target domain, ultimately yielding superior segmentation results.
\section{CONCLUSION}
This paper proposes a SFDA method for semantic segmentation of RSIs, termed ProSFDA. 
To address the limitations of conventional ST methods, 
we introduce prototype-weighted self-training, 
thereby mitigating the negative impact of noisy signals. 
However, due to the weak supervision in noisy regions, 
we propose prototype-contrast feature clustering, which aggregates similar features to enhance the model’s ability to learn from the target domain and improve knowledge transfer. 
Experimental results demonstrate that our method achieves state-of-the-art performance, and ablation studies further confirm the effectiveness of each component in ProSFDA.


%
%

\bibliographystyle{IEEEbib}
\bibliography{strings,refs}

\end{document}